\documentclass{article}
\usepackage{microtype}
\usepackage{graphicx}
\usepackage{booktabs}
\usepackage{xurl}
\usepackage{hyperref}
\usepackage[preprint]{icml2027}
\hypersetup{pdfsubject={Preprint}}
\usepackage{stfloats}
\usepackage{mathtools}
\usepackage{amssymb}
\usepackage{array}
\usepackage{colortbl}
\usepackage{pifont}
\usepackage{tikz}
\usepackage{xspace}

\providecommand{\doi}[1]{doi: \href{https://doi.org/#1}{\nolinkurl{#1}}}
\newcommand{\benchmark}{AVCodeEdit\xspace}
\newcommand{\bpstart}[1]{\par\addvspace{2pt}\noindent\mbox{\textbf{#1.}}\hspace{0.5em}\ignorespaces}

\newcommand{\yes}{\ding{51}}
\newcommand{\cross}{\ding{55}}
\definecolor{ResultsGreen}{HTML}{35840D}
\definecolor{ResultsOrange}{HTML}{ECA138}
\newcommand{\modelentry}[2]{\makebox[11pt][l]{\raisebox{-1.2pt}{#1}}#2}
\definecolor{ProprietaryTint}{HTML}{DCE7EF}
\definecolor{ProprietaryInk}{HTML}{365A73}
\definecolor{OpenWeightTint}{HTML}{E8E0EF}
\definecolor{OpenWeightInk}{HTML}{67507D}
\makeatletter
\newcommand{\modelgroupbar}[4]{%
  \rlap{\smash{\raisebox{-\dp\@arstrutbox}{%
    \begin{tikzpicture}
      \shade[left color=#2,right color=white] (0,0)
        rectangle (#1,\dimexpr\ht\@arstrutbox+\dp\@arstrutbox\relax);
    \end{tikzpicture}}}}%
  \hspace{3pt}\textcolor{#3}{\textbf{\textit{#4}}}%
}
\makeatother
\icmltitlerunning{AVCodeEdit: Benchmarking Coding Agents on Audio-Video Editing}

\begin{document}
\twocolumn[
  \icmltitle{AVCodeEdit: Benchmarking Coding Agents on Audio-Video Editing}
  \icmlsetsymbol{projectlead}{\textdagger}
  \begin{icmlauthorlist}
    \icmlauthor{Zhenghan Chen}{fudan}
    \icmlauthor{Zekai Shao}{fudan,projectlead}
    \icmlauthor{Leting Cheng}{fudan}
    \icmlauthor{Xin Lin}{sysu}
    \icmlauthor{Siming Chen}{fudan}
  \end{icmlauthorlist}
  \icmlaffiliation{fudan}{Fudan University, Shanghai, China}
  \icmlaffiliation{sysu}{Sun Yat-sen University, Guangzhou, China}
  \icmlcorrespondingauthor{Siming Chen}{simingchen@fudan.edu.cn}
  \icmlkeywords{Audiovisual Editing, Coding Agents, Benchmarks}
  \vskip 0.3in
]
\begingroup
  \renewcommand{\today}{November 2026}
  \printAffiliationsAndNotice{\textsuperscript{\textdagger}Project Lead.\enspace}
\endgroup

\begin{abstract}
Code-based audio-video editing offers precise control over visual and audio content, enabling flexible modifications across extended timelines. Advances in multimodal agents have made such editing increasingly practical, requiring the integration of multimodal understanding, instruction following, temporal reasoning, and programmatic reconstruction. 
Yet existing code-based video benchmarks primarily focus on video generation or visual-only reconstruction in specific domains, leaving these combined capabilities insufficiently evaluated in audio-video editing.
We introduce \benchmark{}, a benchmark comprising 1,026 editing tasks with 3,024 editing goals, accompanied by 35,048 fine-grained checklist items. The benchmark spans 11 video categories and 28 subtypes and defines a comprehensive taxonomy of 38 editing operations across joint audio-video, video-only, non-speech audio, and speech editing, covering both local and global modifications across varying video lengths and complexity levels. Our dual-level evaluation combines objective and subjective metrics. Objective evaluation measures video and audio quality, cross-modal correspondence and synchronization, and output validity. Subjective evaluation uses checklists to evaluate edit correctness and the reconstruction fidelity of content outside the requested modifications, alongside five dimensions of audiovisual quality. 
Evaluations of 13 proprietary and open-weight model configurations reveal persistent challenges in edit correctness, reconstruction fidelity, and overall audiovisual quality, particularly for audio and cross-modal requirements.
Code and dataset are released at \url{https://avcodeedit.github.io}.
\end{abstract}

\begin{figure*}[t]
    \centering
    \includegraphics[width=\textwidth]{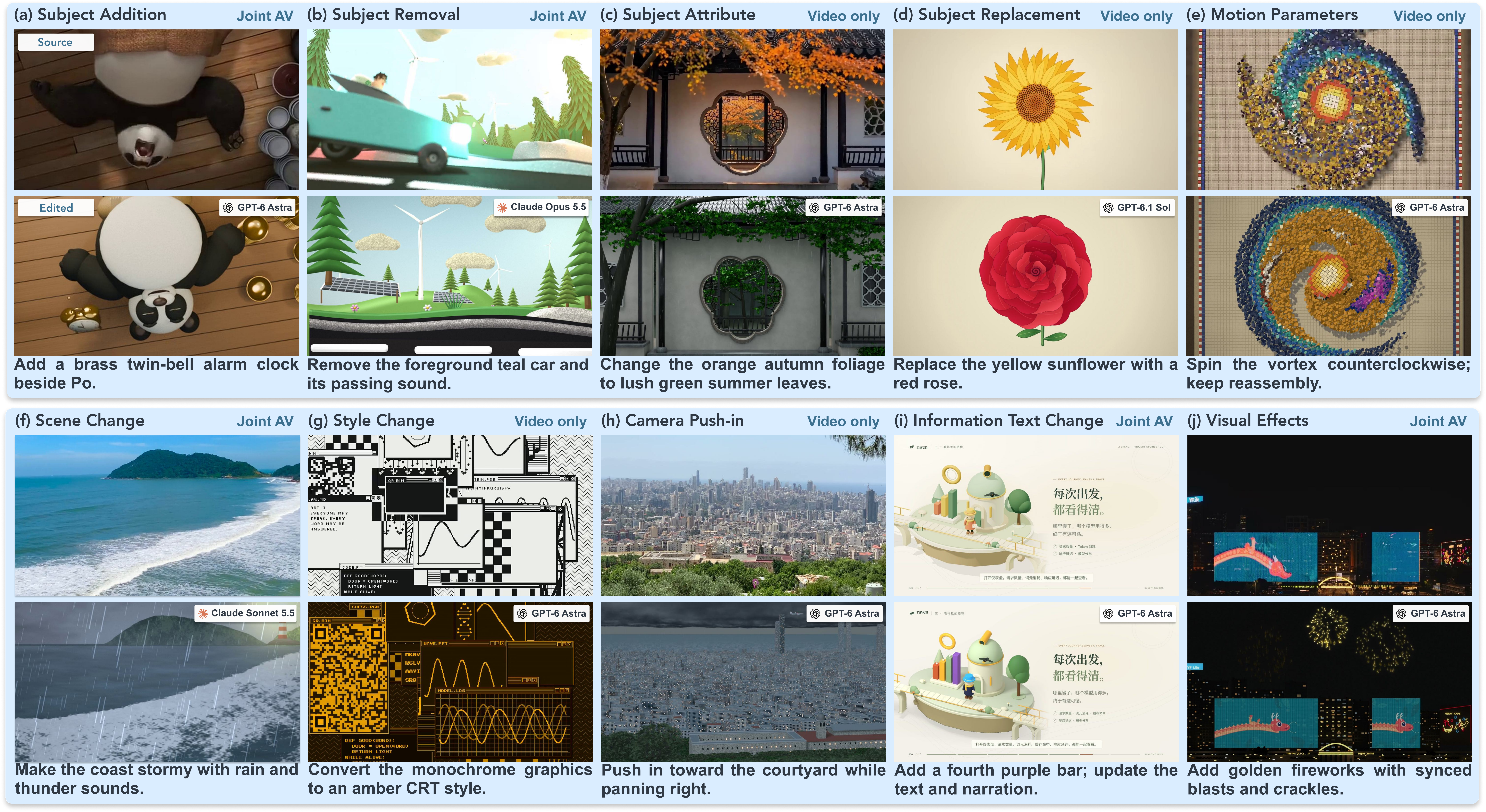}
    \caption{Representative editing tasks in \benchmark{}. Each example pairs a source frame and a shortened editing instruction with a frame from an agent's rendered output. The examples cover visual edits and coordinated audio-video changes.}
    \label{fig:editing_cases}
\vspace{-10pt}
\end{figure*}

\section{Introduction}
\label{sec:introduction}

Advances in multimodal large language models (MLLMs)~\citep{gpt6astra,claudeopus55,gemini4argon} and coding agents~\citep{openai2025codex,anthropic2025claudecode,google2025antigravity} have driven progress in code-based video generation~\citep{c2vcompass,videocoder,sensenova68code2video} and programmatic scene reconstruction from videos~\citep{4dcodebench,worldmirror,grase}. These developments provide a foundation for extending code-based video creation to audio-video editing, where executable programs enable precise and iterative modifications, coordinated visual and audio changes, and editing across extended timelines. Performing such tasks requires agents to integrate audio-video understanding, temporal and compositional reasoning, and program synthesis to implement requested edits while faithfully reconstructing unedited source content.

Prior benchmarks have explored video generation, reconstruction, and editing from different perspectives. Code2Video-Compass~\citep{c2vcompass} benchmarks general-purpose code-based video generation, while VibeVideoBench~\citep{vibevideobench} evaluates audio-video generation. Animation2Code~\citep{animation2code}, BVB~\citep{bvb}, and CodeCap-Bench~\citep{codecap} focus on silent video reconstruction or representation in specific domains, while CoVEBench~\citep{covebench}, OmniEdit-Bench~\citep{omniedit}, and AVE-Compass~\citep{avecompass} evaluate direct generative video editing, mainly on short clips.
Despite these advances, systematic evaluation remains limited for agents that reconstruct both visual and audio content from diverse source videos of varying lengths while implementing multiple editing goals.

To address this gap, we introduce \benchmark{}, a benchmark for code-based audio-video editing in which MLLM agents reconstruct input content and implement multiple editing goals through executable programs that render complete edited videos. \benchmark{} comprises 1,026 editing tasks with 3,024 editing goals across 11 video categories and 28 subtypes, accompanied by 35,048 fine-grained checklist items. 
We establish a taxonomy of 38 editing operations spanning joint audio-video, video-only, non-speech audio, and speech editing, covering local and global modifications across varying video lengths and complexity levels, as illustrated in Fig.~\ref{fig:editing_cases}. We further develop a dual-level evaluation framework combining 10 objective metrics with seven subjective metrics. Objective evaluation measures output validity, video and audio quality, as well as audio-video correspondence and synchronization. 
Subjective evaluation uses instruction following and fidelity preservation checklists to evaluate edit correctness and reconstruction fidelity across visual, audio, and cross-modal aspects, complemented by five quality metrics covering visual presentation, structural integrity, dynamic quality, audio quality, and audio-video timing. We evaluate 13 model configurations using eight agent harnesses to analyze their editing and reconstruction capabilities across editing operations and levels of task complexity.
In summary, our contributions are threefold:
\vspace{-6pt}
\begin{itemize}
    \setlength{\itemsep}{3pt plus 1pt}
    \setlength{\parskip}{0.5pt plus 0.5pt}
    \item We introduce \benchmark{}, which formulates and evaluates agentic code-based audio-video editing through executable reconstruction and editing with multiple goals while preserving unedited content. It comprises 1,026 editing tasks with 3,024 editing goals, covering 38 operations.

    \item We develop a comprehensive dual-level evaluation framework combining objective and subjective metrics to evaluate edit correctness, reconstruction fidelity, and overall audiovisual quality.

    \item We evaluate 13 model configurations across eight agent harnesses, revealing persistent challenges in editing and reconstruction, with lower instruction-following scores for audio and cross-modal requirements.
\end{itemize}

\section{Related Work}
\label{sec:related_work}

\subsection{Code-Based Video Tasks}

Prior work has explored code-based approaches to video generation and reconstruction. Code2Video~\citep{code2video} focuses on educational video generation, while LogoMotion~\citep{logomotion} supports visually grounded logo animation and editing. Code2Video-Compass~\citep{c2vcompass} establishes a benchmark for general-purpose code-based video generation across multiple input modalities, while VibeVideoBench~\citep{vibevideobench} evaluates coding agents on diverse audio-video generation tasks. For video reconstruction, Animation2Code~\citep{animation2code} focuses on web animations, BVB~\citep{bvb} on Blender scenes, 4DCodeBench~\citep{4dcodebench} on scene geometry and dynamics, and CodeCap-Bench~\citep{codecap} on structured 2D videos.
These works provide a foundation for code-based audio-video editing, yet existing benchmarks primarily focus on video generation or domain-specific visual reconstruction, without systematically evaluating audiovisual reconstruction alongside multiple editing goals.

\subsection{Video and Audio-Video Editing}

Existing benchmarks such as VE-Bench~\citep{vebench}, EditBoard~\citep{editboard}, and FiVE-Bench~\citep{fivebench} primarily focus on visual editing, evaluating instruction following, video quality, and content preservation. More recent benchmarks extend evaluation to audio-video editing, with OmniEdit-Bench~\citep{omniedit} incorporating audio editing and AVE-Compass~\citep{avecompass} evaluating coordinated visual and audio modifications. CoVEBench~\citep{covebench} further examines instructions involving multiple editing goals. 
Yet these benchmarks primarily target direct generative editing on short video clips, leaving code-based audio-video editing of longer videos that combines source reconstruction with multiple editing goals insufficiently evaluated.
Tab.~\ref{tab:benchmark_comparison} compares benchmarks.

\begin{table*}[t]
    \centering

\caption{Comparison of benchmarks. Code denotes an executable program requirement. G, R, and E denote generation, reconstruction, and editing. Dual combines objective and subjective metrics. Duration is the average video length in minutes. -- indicates not applicable.}
    \label{tab:benchmark_comparison}
    \begingroup
    \fontsize{8.6pt}{10pt}\selectfont
    \setlength{\tabcolsep}{1.5pt}
    \renewcommand{\arraystretch}{1.03}
    \begin{tabular*}{\textwidth}{@{\extracolsep{\fill}}l r c c l c c c c c@{}}
        \toprule
        \textbf{Benchmark} & \textbf{\#Items} & \textbf{Code} & \textbf{Task} & \textbf{Video types} & \textbf{Audio} & \textbf{\#Edit types} & \textbf{Multiple goals} & \textbf{Evaluation} & \textbf{Duration} \\
        \specialrule{\lightrulewidth}{\aboverulesep}{0pt}

        \rowcolor{black!8}[0pt][0pt]
        \multicolumn{10}{@{}l@{}}{\textit{Code-based generation and reconstruction}} \\
        Code2Video-Compass~\citep{c2vcompass} & 2,104 & \yes & G & CGI, Anim. & \cross & -- & -- & Dual & $>1$ \\
        VibeVideoBench~\citep{vibevideobench} & 239 & \yes & G & CGI, Anim. & \yes & -- & -- & Subjective & $>1$ \\
        BVB~\citep{bvb} & 288 & \yes & R & Real & \cross & -- & -- & Dual & $>1$ \\
        CodeCap-Bench~\citep{codecap} & 323 & \yes & R & Anim. & \cross & -- & -- & Dual & $<1$ \\
        \specialrule{\lightrulewidth}{\aboverulesep}{0pt}

        \rowcolor{black!8}[0pt][0pt]
        \multicolumn{10}{@{}l@{}}{\textit{Direct generative video editing}} \\
        CoVEBench~\citep{covebench} & 626 & \cross & E & Real, CGI & \cross & 19 & \yes & Dual & $<1$ \\
        FiVE-Bench~\citep{fivebench} & 420 & \cross & E & Real, CGI & \cross & 6 & \cross & Objective & $<1$ \\
        OmniEdit-Bench~\citep{omniedit} & 790 & \cross & E & Real, CGI & \yes & 79 & \yes & Subjective & $<1$ \\
        AVE-Compass~\citep{avecompass} & 196 & \cross & E & Real, CGI, Anim. & \yes & 28 & \cross & Dual & $<1$ \\
        \specialrule{\lightrulewidth}{\aboverulesep}{0pt}

        \rowcolor{black!8}[0pt][0pt]
        \multicolumn{10}{@{}l@{}}{\textit{Code-based audio-video editing}} \\
        \textbf{\benchmark{} (ours)} & \textbf{1,026} & \yes & \textbf{R+E} & Real, CGI, Anim. & \yes & \textbf{38} & \yes & \textbf{Dual} & $\mathbf{>1}$ \\
        \bottomrule
    \end{tabular*}
    \endgroup
\vspace{-10pt}
\end{table*}

\section{AVCodeEdit Benchmark}
\label{sec:benchmark}

\subsection{Task Formulation}
\label{sec:task}

\subsubsection{Task Definition}
\label{sec:task_definition}

\benchmark evaluates models on code-based audio-video editing using their corresponding agent harnesses. Each task provides a source video with its original soundtrack and a natural-language instruction specifying one or more compatible editing goals. Each goal identifies a target, a requested change, and its temporal scope. The agent must understand the source's visual and auditory content and ground each editing goal in the relevant subjects, events, and time intervals. It must deliver an executable program and its complete rendered audio-video output, implementing the requested changes and their necessary visual or acoustic consequences while faithfully reconstructing the remaining source content.

All agents are evaluated in a standardized execution environment under identical computational resource constraints. Each agent can inspect the source video, write and execute code, and review renders. Visuals and non-speech audio must be generated programmatically without source-media reuse, while TTS is permitted to produce intelligible speech.

\subsubsection{Taxonomy and Complexity}
\label{sec:taxonomy}

\begin{figure}[t]
    \centering

    \includegraphics[width=0.8\columnwidth]{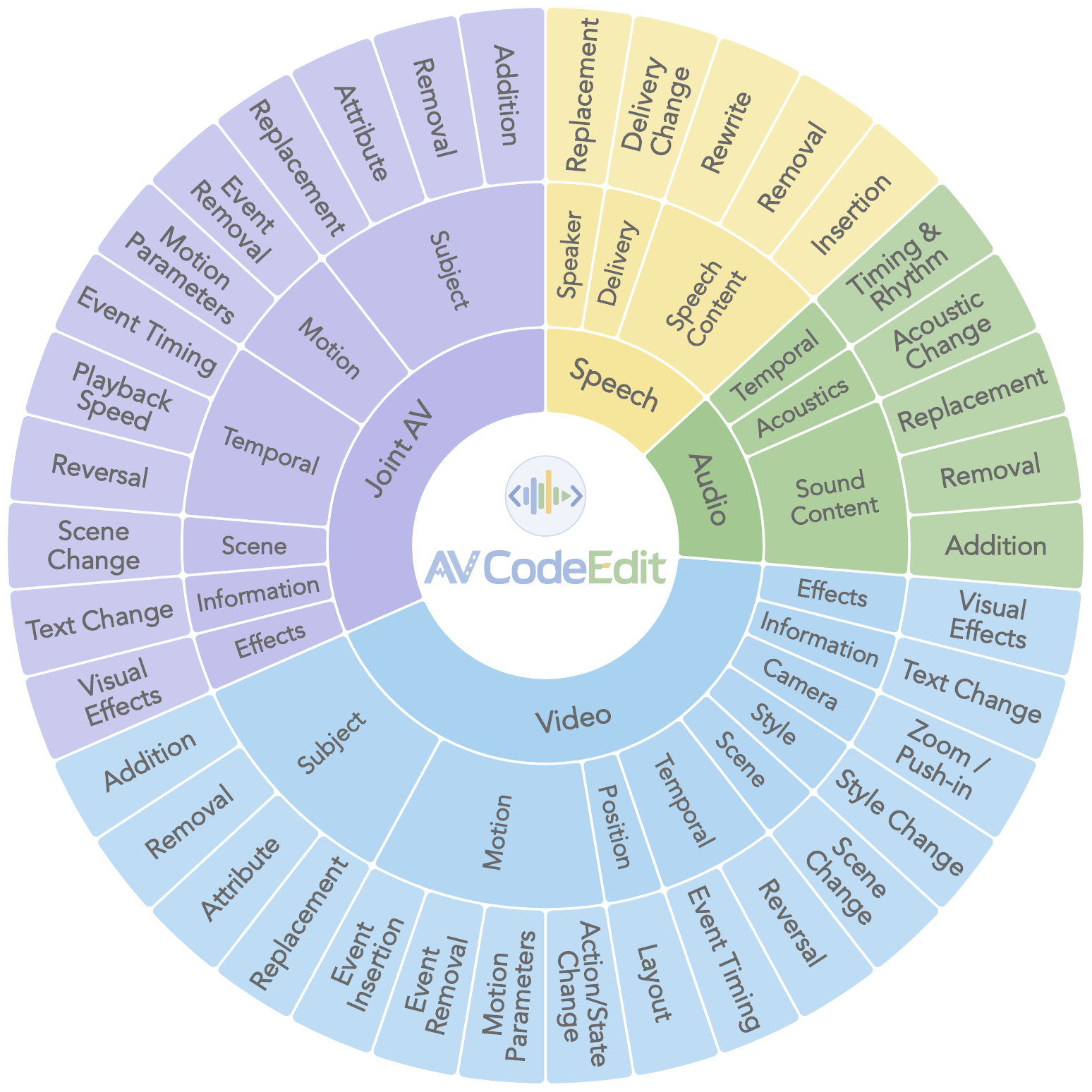}
    \caption{Editing taxonomy with 38 operations across joint audio-video, video-only, non-speech audio, and speech editing.}
    \label{fig:edit_taxonomy}
\vspace{-16pt}
\end{figure}

\bpstart{Editing Taxonomy}
Fig.~\ref{fig:edit_taxonomy} shows our three-level taxonomy of four modality groups, 14 editing aspects, and 38 operations. The modality groups are \textit{joint audio-video}, \textit{video-only}, \textit{non-speech audio}, and \textit{speech} editing, containing 12, 16, 5, and 5 operations. Joint audio-video editing requires coordinated visual and acoustic changes within a single goal, while the other groups target their respective content.
To reflect the fine-grained control required by code-based editing, the taxonomy covers subject and scene modifications, motion and temporal changes, visual effects, text editing, and audio and speech modifications. It further distinguishes motion parameter changes, event timing, playback speed adjustments, and reversal, which involve different temporal scopes. Each editing goal is assigned to one operation, and an instruction may combine compatible goals within or across groups.

\bpstart{Task Complexity}
Code-based editing requires both reconstructing the source and implementing the requested modifications. We therefore annotate reconstruction and editing complexity separately. Each source video is assessed along five reconstruction dimensions, each with three levels. \textit{Construction Complexity} captures scene geometry and visual composition. \textit{Dynamic Animation Complexity} captures motion, camera behavior, and coordinated visual changes. \textit{State-Dependency Complexity} captures temporal dependencies among objects, events, and states. \textit{Audio Reconstruction Complexity} covers sound synthesis and organization, while \textit{Audio-Video Complexity} captures event--sound associations and temporal coordination. For editing, we characterize task difficulty along three attributes. \textit{Object Localization} (Levels~1--2) captures the difficulty of identifying the intended editing target. \textit{Audio Source Complexity} (Levels~1--3) characterizes the audio sources involved in the modifications. \textit{Cross-Modal Linkage} (Levels~1--2) indicates whether the relationships between visual and audio elements are explicitly specified in the instruction or must be inferred.

\subsection{Data Construction}
\label{sec:data_construction}

\begin{figure*}[t]

    \centering
    \includegraphics[width=\textwidth]{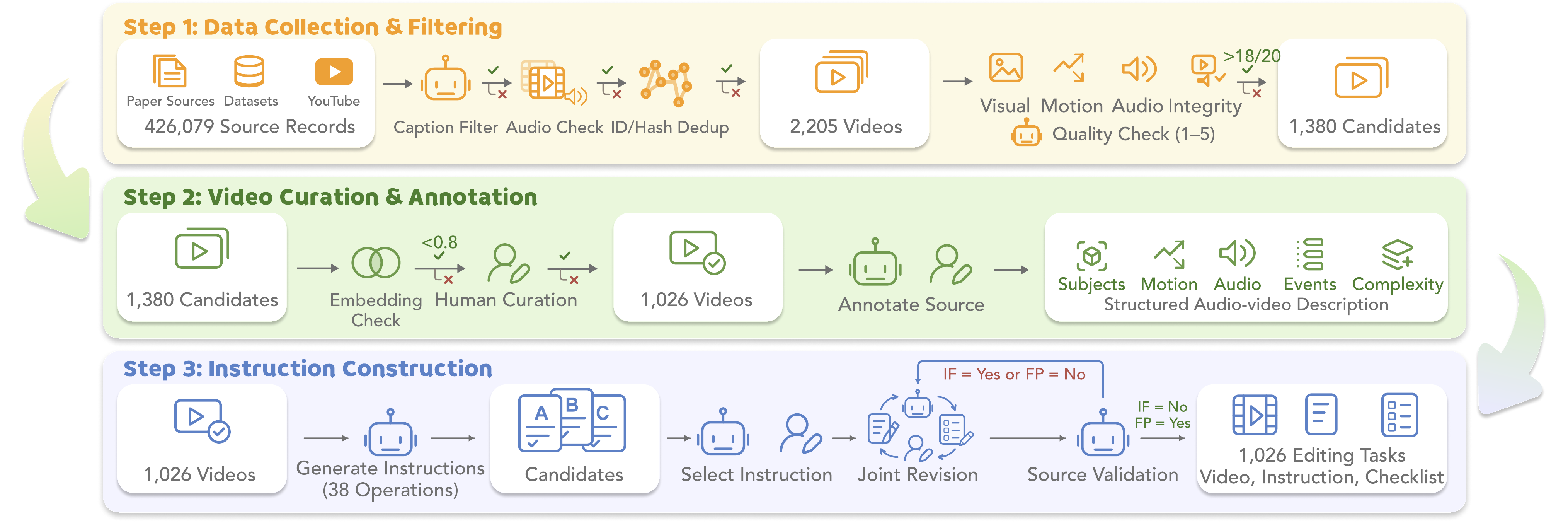}
    \caption{Overview of the \benchmark data construction pipeline. We collect, screen, and curate source videos, annotate audiovisual events, and construct editing instructions with source-grounded instruction-following and fidelity-preserving checklists.}
    \label{fig:data_construction}
\vspace{-12pt}
\end{figure*}

Three co-authors contribute to source collection and screening, video curation and annotation, and editing instruction construction. Fig.~\ref{fig:data_construction} summarizes the workflow.

\bpstart{Source collection and screening}
We compile 426,079 source records from benchmark sources, including Code2Video-Compass~\citep{c2vcompass}, AVE-Compass~\citep{avecompass}, OmniEdit-Bench~\citep{omniedit}, and CMD-AM~\citep{cmdam}, video datasets such as VPData~\citep{videopainter} and Vript~\citep{vript}, and curated YouTube channels and topic searches, supplemented by exploratory sources. We use GPT-6.1 Sol~\citep{gpt61sol} to shortlist videos based on their captions and descriptions, with an emphasis on diversity in scenes, subjects, and actions. The shortlisted videos are then checked for decodable, non-silent audio tracks and filtered for duplicates using URLs, video IDs, and file hashes, resulting in 2,205 videos. 
We employ Gemini 3.8 Flash~\citep{gemini38flash} to assess the videos across four dimensions, each rated on a five-point scale. Visual quality evaluates clarity and legibility, motion quality assesses the continuity of subject and camera movements, and audio quality considers intelligibility and distortion. Content integrity examines the completeness of visual and auditory events and their temporal context. Videos scoring above 18 out of 20 are selected, yielding 1,380 candidates.

\bpstart{Curation and annotation}
To further ensure diversity, we compute pairwise embedding similarities among the 1,380 remaining videos and manually review pairs with similarity scores of at least 0.8. Human reviewers also examine the full clips with their original soundtracks to verify that relevant objects, events, and sounds are fully captured. This process yields 1,026 curated videos spanning diverse content, visual styles, audio characteristics, and reconstruction complexity.

For each selected video, we employ Gemini 3.8 Flash~\citep{gemini38flash} to generate initial annotations, which are subsequently reviewed and refined by human annotators. These annotations capture key subjects, camera and object movements, and the temporal progression of visual and audio events. We further annotate video content, visual style, and the five reconstruction complexity dimensions defined in Sec.~\ref{sec:taxonomy}.

\bpstart{Instruction construction}
For each source video, we use Gemini 3.8 Flash~\citep{gemini38flash} to generate two to three candidate editing instructions based on our taxonomy of 38 operations. Each instruction contains one or more editing goals involving local or global modifications. Human reviewers select and refine the most suitable instruction with the assistance of GPT-6.1 Sol~\citep{gpt61sol}, considering its consistency with the source video, compatibility among editing goals, and verifiability of the requested modifications. Each finalized task is annotated along three editing complexity attributes, including object localization, audio source complexity, and cross-modal linkage (Sec.~\ref{sec:taxonomy}).

Given the source video and finalized editing instruction, we use Gemini 3.8 Flash~\citep{gemini38flash} to generate binary instruction-following and fidelity-preserving questions guided by a taxonomy of nine primary dimensions and 22 sub-dimensions. The primary dimensions are as follows. (1) \textit{Content} examines text and information. (2) \textit{World Knowledge} assesses factual knowledge. (3) \textit{Attribute} evaluates appearance and quantity. (4) \textit{Cinematography} covers lighting, framing, and color grading. (5) \textit{Relations} examines spatial and logical relationships. (6) \textit{Dynamics} assesses motion, interaction, transformation, camera motion, and temporal relations. (7) \textit{Aesthetics} evaluates style and mood. (8) \textit{Sound} covers sound effects, speech, absence of speech, and music. (9) \textit{Audiovisual} assesses correspondence between visual and auditory events. Each question is also labeled as visual question answering (VQA), audio question answering (AQA), or audiovisual question answering (AVQA) based on the evidence required. Human reviewers, assisted by GPT-6.1 Sol~\citep{gpt61sol}, review and refine the generated questions to ensure comprehensive coverage, eliminate redundancy, and maintain consistency with the source video and finalized instruction.
Finally, Gemini 3.8 Flash~\citep{gemini38flash} validates the checklists against the source video, expecting preservation conditions to hold and editing requirements to remain unmet. Human reviewers resolve discrepancies by revising instructions or checklists.

\subsection{Dataset Statistics}
\label{sec:dataset_statistics}
\begin{figure*}[t]

    \centering
    \includegraphics[width=\textwidth]{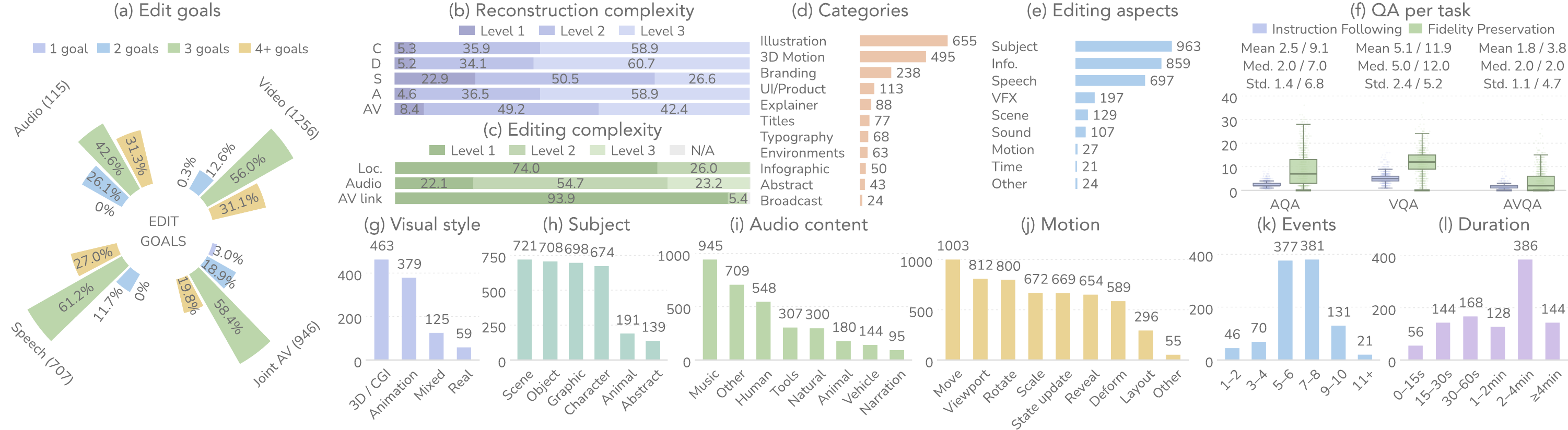}
\caption{Dataset composition of \benchmark. The top row shows task statistics, including editing goals, complexity, categories, and QA statistics. In (b), C, D, S, A, and AV denote construction, dynamic animation, state dependency, audio reconstruction, and audio-video complexity, respectively. The bottom row shows source video diversity across styles, subjects, audio, motion, events, and duration.}
    \label{fig:data_statistics}
\vspace{-15pt}
\end{figure*}

\benchmark comprises 1,026 editing tasks, each with a source video and an instruction, yielding 3,024 editing goals across 38 operations and an average of 2.95 goals per task. These goals include 946 joint audio-video edits, 1,256 video-only edits, 115 non-speech audio edits, and 707 speech edits. The dataset includes 35,048 binary checklist questions, of which 9,619 evaluate instruction following and 25,429 evaluate fidelity preservation. Fig.~\ref{fig:data_statistics} provides an overview of the dataset composition and diversity.

\subsection{Dual-Level Evaluation Framework}
\label{sec:evaluation}

We evaluate generated videos through complementary objective and subjective evaluations. Specifically, 10 objective metrics measure video quality, audio quality, cross-modal quality, and output validity, while we employ Gemini 4 Argon to evaluate instruction following, fidelity preservation, and five dimensions of audiovisual quality.

\subsubsection{Objective Evaluation}
\label{sec:objective}

\bpstart{Output validity}
\textbf{Valid Output Rate (VOR)} measures the proportion of tasks producing decodable videos that meet output specifications and comply with content and tool-use rules. Invalid outputs receive zero scores across all evaluation metrics to prevent inflated averages from excluding failed tasks or rewarding prohibited source reuse.

\bpstart{Video quality}
We evaluate the edited video in terms of visual quality, appearance stability, and temporal coherence. \textbf{Technical Quality (TQ)} measures visual clarity and artifacts using the technical branch of DOVER++~\citep{dover}, while \textbf{Visual Aesthetics (VA)} evaluates visual appeal using Aesthetic Predictor V2.5~\citep{aesthetic}. \textbf{Motion Smoothness (MS)} measures motion continuity based on frame interpolation errors from AMT~\citep{amt}, while \textbf{Appearance Stability (AS)} evaluates visual consistency through DINOv3~\citep{dinov3} feature similarities between adjacent frames. Both metrics exclude comparisons across cuts found by PySceneDetect~\citep{pyscenedetect}.

\bpstart{Audio quality}
We evaluate the generated soundtrack in terms of production quality, content usefulness, and speech naturalness. \textbf{Production Quality (PQ)} measures audio clarity and technical fidelity, while \textbf{Content Usefulness (CU)} evaluates the suitability of audio for content creation, both using Audiobox Aesthetics~\citep{audioboxaesthetics}. 
\textbf{Speech Quality (SQ)} measures speech naturalness using UTMOSv2~\citep{utmosv2} on tasks expected to contain speech. If expected speech is undetected, outputs score zero.

\bpstart{Cross-Modal quality}
We evaluate the semantic correspondence and temporal synchronization between visual and audio content. \textbf{Audiovisual Correspondence (AVC)} measures semantic consistency using ImageBind~\citep{imagebind} feature similarities across aligned segments, with duration-weighted averaging. \textbf{Audiovisual Synchronization (AVS)} evaluates the temporal alignment of visible events and corresponding sounds using SyncReward Stage~2~\citep{syncreward}, averaging scores across overlapping temporal windows.

\subsubsection{Subjective Evaluation}
\label{sec:subjective}

\bpstart{Instruction Following and Fidelity Preservation}
Using the QA checklists from Sec.~\ref{sec:data_construction}, the MLLM judge evaluates edit correctness (IF) and reconstruction fidelity for content that should be preserved (FP), including unchanged attributes of edited objects. Checklist items receive binary scores of 1 when fully satisfied and 0 otherwise. For each task, we compute IF and FP as the percentages of satisfied items in their respective checklists, then average task scores across the evaluation subset. We also report IF and FP separately for VQA, AQA, and AVQA, averaging over tasks with applicable questions in each modality.

\bpstart{Audiovisual quality} Overall audiovisual quality is the unweighted mean of five 1--5 ratings. \textbf{Visual Presentation (VP)} evaluates visual composition, typography, clarity, and overall presentation quality. \textbf{Structural Integrity (SI)} evaluates the consistency of subject appearance, structure, identity, and state within shots and across shots with established continuity. \textbf{Dynamic Quality (DQ)} evaluates subject and camera motion, transitions, and the physical plausibility of interactions in the depicted scene. \textbf{Audio Quality (AQ)} measures sound clarity, mixing quality, timbre, and acoustic consistency. \textbf{Audio-Video Timing (AVT)} evaluates synchronization between visual events and corresponding sounds, including the timing and continuity of music and ambience.

\section{Experiments}
\label{sec:experiments}
\begin{table*}[t]
\centering

\caption{Results on AVCodeEdit. Overall denotes overall audiovisual quality (mean of VP, SI, DQ, AQ, and AVT). Within each group, \textbf{bold} and \underline{underlined} mark the best and second-best unrounded scores. Rows are sorted by Overall. Baseline is the source-video reference.}
\label{tab:main_results}
\begingroup
\fontsize{7.5pt}{9pt}\selectfont
\setlength{\tabcolsep}{0.85pt}
\renewcommand{\arraystretch}{1.03}
\newcommand{\baselinecell}[1]{\raisebox{-0.7pt}{#1}}
\begin{tabular*}{\textwidth}{@{\extracolsep{\fill}}l*{18}{c}@{}}
\toprule
\textbf{Model}
& \multicolumn{1}{c}{\textbf{Validity}}
& \multicolumn{4}{c}{\textbf{Video Quality}}
& \multicolumn{3}{c}{\textbf{Audio Quality}}
& \multicolumn{2}{c}{\textbf{AV Quality}}
& \multicolumn{2}{c}{\shortstack{\textbf{Task}\\\textbf{Completion}}}
& \multicolumn{6}{c}{\textbf{Audiovisual Quality}} \\
\cmidrule(lr){2-2}\cmidrule(lr){3-6}\cmidrule(lr){7-9}
\cmidrule(lr){10-11}\cmidrule(lr){12-13}\cmidrule(l){14-19}
& \shortstack{\textbf{VOR}$\uparrow$\\\textbf{(0--1)}}
& \textbf{TQ}$\uparrow$ & \textbf{VA}$\uparrow$ & \textbf{MS}$\uparrow$ & \textbf{AS}$\uparrow$
& \textbf{PQ}$\uparrow$ & \textbf{CU}$\uparrow$ & \textbf{SQ}$\uparrow$
& \textbf{AVC}$\uparrow$ & \textbf{AVS}$\uparrow$
& \shortstack{\textbf{IF}$\uparrow$\\\textbf{(\%)}} & \shortstack{\textbf{FP}$\uparrow$\\\textbf{(\%)}}
& \textbf{VP}$\uparrow$ & \textbf{SI}$\uparrow$ & \textbf{DQ}$\uparrow$ & \textbf{AQ}$\uparrow$ & \textbf{AVT}$\uparrow$ & \textbf{Overall}$\uparrow$ \\
\specialrule{\lightrulewidth}{\aboverulesep}{0pt}
\rlap{\smash{\raisebox{-\dp\csname @arstrutbox\endcsname}{%
  \textcolor{black!8}{\rule{\textwidth}{%
    \dimexpr\ht\csname @arstrutbox\endcsname+\dp\csname @arstrutbox\endcsname\relax}}}}}%
\baselinecell{\textbf{Baseline}} & {--}
& \baselinecell{0.115} & \baselinecell{3.813} & \baselinecell{0.978} & \baselinecell{0.898}
& \baselinecell{7.189} & \baselinecell{6.726} & \baselinecell{2.201}
& \baselinecell{0.213} & \baselinecell{0.662}
& \baselinecell{0.05} & \baselinecell{99.83}
& \baselinecell{4.11} & \baselinecell{4.15} & \baselinecell{4.13} & \baselinecell{4.08} & \baselinecell{4.13} & \baselinecell{4.12} \\
\specialrule{\lightrulewidth}{\aboverulesep}{0pt}
\multicolumn{19}{@{}l@{}}{\modelgroupbar{\textwidth}{ProprietaryTint}{ProprietaryInk}{Proprietary models}} \\
\modelentry{\includegraphics[width=7.5pt,height=7.5pt,keepaspectratio]{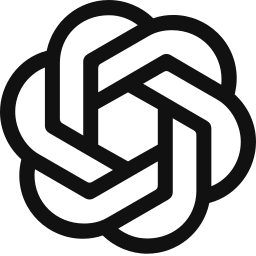}}{GPT-6 Astra} & \textbf{1.000} & \textbf{0.125} & \underline{3.997} & \underline{0.989} & 0.948 & \underline{7.258} & 6.904 & 2.209 & \underline{0.176} & \textbf{0.585} & 83.22 & \textbf{58.60} & \textbf{2.87} & \textbf{3.15} & \textbf{2.97} & \underline{2.91} & \underline{3.02} & \textbf{2.99} \\
\modelentry{\includegraphics[width=7.5pt,height=7.5pt,keepaspectratio]{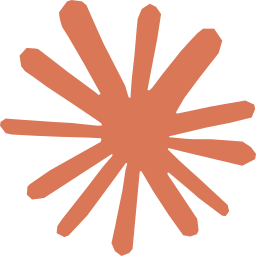}}{Claude Opus 5.5} & \textbf{1.000} & 0.115 & 3.865 & 0.987 & 0.938 & \textbf{7.340} & \textbf{7.040} & 2.163 & 0.167 & 0.552 & 83.20 & \underline{57.43} & \underline{2.79} & \underline{3.05} & \underline{2.95} & \textbf{2.95} & \textbf{3.07} & \underline{2.96} \\
\modelentry{\includegraphics[width=7.5pt,height=7.5pt,keepaspectratio]{figures/logos/anthropic.png}}{Claude Sonnet 5.5} & 0.985 & 0.114 & 3.742 & 0.973 & 0.923 & 7.075 & 6.776 & 1.928 & 0.169 & \underline{0.584} & \underline{83.38} & 54.47 & 2.69 & 2.87 & 2.83 & 2.81 & 2.93 & 2.83 \\
\modelentry{\includegraphics[width=7.5pt,height=7.5pt,keepaspectratio]{figures/logos/openai.png}}{GPT-6.1 Sol} & 0.985 & \underline{0.122} & 3.888 & 0.975 & 0.937 & 7.116 & 6.782 & 2.133 & 0.172 & 0.534 & 79.99 & 50.14 & 2.70 & 2.94 & 2.81 & 2.80 & 2.87 & 2.82 \\
\modelentry{\includegraphics[width=7.5pt,height=7.5pt,keepaspectratio]{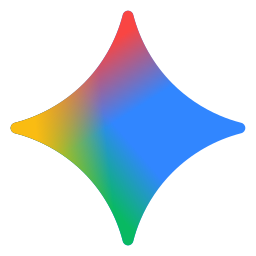}}{Gemini 4 Argon} & 0.961 & 0.120 & 3.622 & 0.948 & 0.905 & 7.146 & 6.763 & 2.145 & \textbf{0.181} & 0.494 & \textbf{86.73} & 57.00 & 2.59 & 2.68 & 2.61 & 2.60 & 2.89 & 2.67 \\
\modelentry{\includegraphics[width=7.5pt,height=7.5pt,keepaspectratio]{figures/logos/openai.png}}{GPT-6.1 Terra} & 0.990 & 0.118 & 3.888 & 0.983 & \underline{0.949} & 7.228 & \underline{6.963} & \textbf{2.330} & 0.169 & 0.493 & 79.55 & 42.66 & 2.51 & 2.96 & 2.60 & 2.57 & 2.69 & 2.67 \\
\modelentry{\includegraphics[width=7.5pt,height=7.5pt,keepaspectratio]{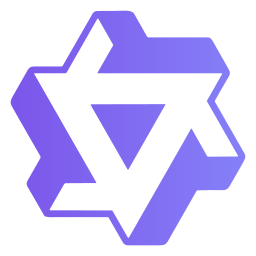}}{Qwen3.8-Flash} & \underline{0.995} & 0.108 & 3.769 & 0.983 & 0.942 & 6.858 & 6.504 & 1.482 & 0.167 & 0.567 & 70.77 & 35.00 & 2.24 & 2.55 & 2.47 & 2.43 & 2.59 & 2.46 \\
\modelentry{\includegraphics[width=7.5pt,height=7.5pt,keepaspectratio]{figures/logos/openai.png}}{GPT-6 Luna} & \textbf{1.000} & 0.115 & \textbf{4.010} & \textbf{0.994} & \textbf{0.975} & 6.751 & 6.403 & \underline{2.261} & 0.161 & 0.474 & 60.40 & 23.01 & 2.16 & 2.60 & 2.30 & 2.30 & 2.35 & 2.34 \\
\modelentry{\includegraphics[width=7.5pt,height=7.5pt,keepaspectratio]{figures/logos/anthropic.png}}{Claude Haiku 5.5} & 0.849 & 0.096 & 3.272 & 0.842 & 0.812 & 5.814 & 5.532 & 1.736 & 0.134 & 0.436 & 58.47 & 26.25 & 1.95 & 2.33 & 2.11 & 2.00 & 2.19 & 2.12 \\
\specialrule{\lightrulewidth}{\aboverulesep}{0pt}
\multicolumn{19}{@{}l@{}}{\modelgroupbar{\textwidth}{OpenWeightTint}{OpenWeightInk}{Open-weight models}} \\
\modelentry{\includegraphics[width=7.5pt,height=7.5pt,keepaspectratio]{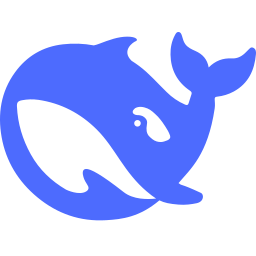}}{DeepSeek-V4.1-Flash} & \textbf{0.990} & \textbf{0.119} & \textbf{3.926} & \textbf{0.980} & \textbf{0.948} & \textbf{7.044} & \textbf{6.778} & 1.682 & \textbf{0.162} & \textbf{0.527} & \textbf{73.73} & \textbf{42.13} & \textbf{2.59} & \textbf{2.85} & \textbf{2.73} & \textbf{2.72} & \textbf{2.81} & \textbf{2.74} \\
\modelentry{\includegraphics[width=7.5pt,height=7.5pt,keepaspectratio]{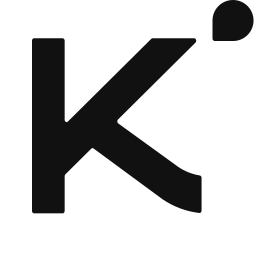}}{Kimi K3.1 Pro} & \underline{0.966} & \underline{0.110} & \underline{3.764} & \underline{0.957} & \underline{0.925} & \underline{6.969} & \underline{6.633} & \textbf{2.247} & \underline{0.161} & \underline{0.473} & \underline{64.72} & \underline{27.25} & \underline{2.22} & \underline{2.40} & \underline{2.30} & \underline{2.51} & \underline{2.47} & \underline{2.38} \\
\modelentry{\includegraphics[width=7.5pt,height=7.5pt,keepaspectratio]{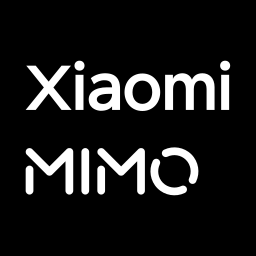}}{MiMo-V2.6-Pro} & 0.927 & 0.103 & 3.499 & 0.920 & 0.890 & 6.392 & 6.001 & \underline{2.042} & 0.156 & 0.422 & 53.31 & 20.84 & 1.99 & 2.36 & 2.17 & 2.18 & 2.20 & 2.18 \\
\modelentry{\includegraphics[width=7.5pt,height=7.5pt,keepaspectratio]{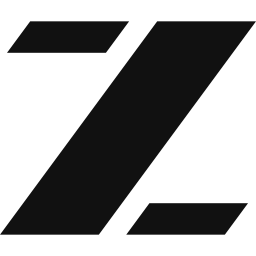}}{GLM-5.3-Flash} & 0.576 & 0.066 & 2.156 & 0.569 & 0.543 & 4.050 & 3.854 & 1.041 & 0.104 & 0.310 & 38.48 & 20.93 & 1.29 & 1.47 & 1.42 & 1.44 & 1.51 & 1.43 \\
\bottomrule
\end{tabular*}

\endgroup
\vspace{-13pt}
\end{table*}

\subsection{Experimental Setup}
\label{sec:experiments_setup}
\bpstart{Models and agents}
We evaluate 13 proprietary and open-weight MLLM configurations across eight model families. The evaluated models include GPT-6 Astra, GPT-6.1 Sol, GPT-6.1 Terra, and GPT-6 Luna~\citep{gpt6astra,gpt61sol,gpt61terra,gpt6luna}, Claude Opus 5.5, Claude Sonnet 5.5, and Claude Haiku 5.5~\citep{claudeopus55,claudesonnet55,claudehaiku55}, Gemini 4 Argon~\citep{gemini4argon}, Kimi K3.1 Pro~\citep{kimik31pro}, Qwen3.8-Flash~\citep{qwen38flash}, GLM-5.3-Flash~\citep{glm53flash}, DeepSeek-V4.1-Flash-1017~\citep{deepseekv41flash}, and MiMo-V2.6-Pro~\citep{mimov26pro}.
We evaluate models using their corresponding agent harnesses, including Codex~\citep{openai2025codex}, Claude Code~\citep{anthropic2025claudecode}, Antigravity CLI~\citep{antigravitycli}, Kimi Code CLI~\citep{kimicode}, Qwen Code~\citep{qwencode}, ZCode~\citep{zcode}, DeepSeek Harness~\citep{deepseekharness}, and MiMo Code~\citep{mimocode}.

\bpstart{Experimental settings}
We select 205 \mbox{\benchmark{}} tasks to form a fixed evaluation subset. Compared with the remaining tasks, this subset shows no statistically significant differences across the five reconstruction complexity dimensions, object localization, audio source complexity, and editing categories (Pearson's $\chi^2$ tests, $p_{\mathrm{Holm}}>0.05$, $V<0.10$). Instruction lengths are also comparable (Welch's $t$-test, $p_{\mathrm{Holm}}>0.05$, $|g|<0.1$), with Holm correction applied jointly across all nine tests. We also evaluate the original source videos from these 205 tasks using the same metrics as a reference baseline.
All agents use their respective models' default reasoning settings and run in isolated workspaces with standardized computing resources and access to preinstalled tools such as Blender~\citep{blender}, Manim~\citep{manim}, Three.js~\citep{threejs}, Remotion~\citep{remotion}, FFmpeg~\citep{ffmpeg}, and Piper TTS~\citep{piper}.
Each run allows 90 minutes for code generation and refinement and unlimited tool calls. Agents can inspect source videos, execute code, and refine their implementations using render feedback.

\bpstart{Evaluation metrics}
We evaluate all model outputs using the dual-level evaluation framework described in Sec.~\ref{sec:evaluation}.
Invalid outputs receive zero content scores, while fully silent outputs receive zero scores on audio and cross-modal quality metrics, as all tasks require sound.

\subsection{Main Results}

Tab.~\ref{tab:main_results} reports the evaluation results of all tested models.

\bpstart{Performance across metrics}
Among the models, GPT-6 Astra achieves the highest overall audiovisual quality score of 2.99, leading in visual presentation, structural integrity, dynamic quality, and fidelity preservation. 
Claude Opus 5.5 follows at 2.96, leading in audio-related metrics PQ, CU, AQ, and AVT. Gemini 4 Argon achieves the highest IF of 86.73\%, while DeepSeek-V4.1-Flash leads open-weight models in overall audiovisual quality, IF, and FP.

\bpstart{Audio-video editing remains challenging}
The highest IF and FP scores reach 86.73\% and 58.60\%, respectively, while all reported models obtain overall audiovisual quality scores below 3.0. Compared with the original source videos used as a reference baseline, for which manual inspection confirms that FP errors stem from MLLM judge misclassifications, the highest FP score is 41.23 percentage points lower, and the highest overall audiovisual quality score is 1.13 points lower. These performance gaps reflect the challenges of accurately implementing requested edits, faithfully reconstructing unmodified content, and maintaining audiovisual quality. Sec.~\ref{sec:modality_analysis} examines these challenges by modality.

\bpstart{Performance discrepancies}
Model performance varies across objective and subjective metrics. For example, GPT-6 Luna leads on several objective metrics but scores 2.34 in overall audiovisual quality and 23.01\% in FP, indicating that strong objective scores do not necessarily reflect accurate editing, faithful reconstruction, or high audiovisual quality. 
This demonstrates the value of dual-level evaluation.

\section{Further Analysis}
\label{sec:further_analysis}

We further analyze modality performance, efficiency, task complexity, editing robustness, and judge reliability.

\subsection{Modality Analysis}
\label{sec:modality_analysis}
\begin{table}[t]
\centering

\caption{IF/FP (\%) by modality. Per-group marks follow Tab.~\ref{tab:main_results}.}
\label{tab:modality_results}
\begingroup
\footnotesize
\setlength{\tabcolsep}{1.6pt}
\renewcommand{\arraystretch}{1.0}
\begin{tabular*}{\columnwidth}{@{\extracolsep{\fill}}l*{6}{c}@{}}
\toprule
\textbf{Model}
& \multicolumn{2}{c}{\textbf{VQA}}
& \multicolumn{2}{c}{\textbf{AQA}}
& \multicolumn{2}{c}{\textbf{AVQA}} \\
\cmidrule(lr){2-3}\cmidrule(lr){4-5}\cmidrule(lr){6-7}
& \textbf{IF}$\uparrow$ & \textbf{FP}$\uparrow$ & \textbf{IF}$\uparrow$ & \textbf{FP}$\uparrow$ & \textbf{IF}$\uparrow$ & \textbf{FP}$\uparrow$ \\
\specialrule{\lightrulewidth}{\aboverulesep}{0pt}
\multicolumn{7}{@{}l@{}}{\modelgroupbar{\columnwidth}{ProprietaryTint}{ProprietaryInk}{Proprietary models}} \\
\modelentry{\includegraphics[width=7.5pt,height=7.5pt,keepaspectratio]{figures/logos/openai.png}}{GPT-6 Astra} & \underline{89.17} & \textbf{64.55} & 79.76 & 53.56 & 71.68 & \underline{42.01} \\
\modelentry{\includegraphics[width=7.5pt,height=7.5pt,keepaspectratio]{figures/logos/anthropic.png}}{Claude Opus 5.5} & 86.31 & \underline{61.59} & 82.87 & \underline{55.59} & 76.07 & \textbf{42.70} \\
\modelentry{\includegraphics[width=7.5pt,height=7.5pt,keepaspectratio]{figures/logos/anthropic.png}}{Claude Sonnet 5.5} & 85.58 & 58.48 & \textbf{84.63} & 52.38 & \underline{76.15} & 35.66 \\
\modelentry{\includegraphics[width=7.5pt,height=7.5pt,keepaspectratio]{figures/logos/openai.png}}{GPT-6.1 Sol} & 84.47 & 55.89 & 77.61 & 48.41 & 70.07 & 28.95 \\
\modelentry{\includegraphics[width=7.5pt,height=7.5pt,keepaspectratio]{figures/logos/gemini.png}}{Gemini 4 Argon} & \textbf{91.36} & 59.01 & \underline{84.59} & \textbf{55.78} & \textbf{79.89} & 39.59 \\
\modelentry{\includegraphics[width=7.5pt,height=7.5pt,keepaspectratio]{figures/logos/openai.png}}{GPT-6.1 Terra} & 85.29 & 50.04 & 76.25 & 37.10 & 70.39 & 22.76 \\
\modelentry{\includegraphics[width=7.5pt,height=7.5pt,keepaspectratio]{figures/logos/qwen.png}}{Qwen3.8-Flash} & 77.78 & 37.64 & 64.02 & 37.06 & 56.81 & 20.16 \\
\modelentry{\includegraphics[width=7.5pt,height=7.5pt,keepaspectratio]{figures/logos/openai.png}}{GPT-6 Luna} & 66.19 & 24.49 & 59.51 & 26.49 & 42.90 & 8.86 \\
\modelentry{\includegraphics[width=7.5pt,height=7.5pt,keepaspectratio]{figures/logos/anthropic.png}}{Claude Haiku 5.5} & 61.51 & 28.27 & 57.03 & 26.04 & 46.26 & 12.19 \\
\specialrule{\lightrulewidth}{\aboverulesep}{0pt}
\multicolumn{7}{@{}l@{}}{\modelgroupbar{\columnwidth}{OpenWeightTint}{OpenWeightInk}{Open-weight models}} \\
\modelentry{\includegraphics[width=7.5pt,height=7.5pt,keepaspectratio]{figures/logos/deepseek.png}}{DeepSeek-V4.1-Flash} & \textbf{80.38} & \textbf{45.24} & \textbf{69.33} & \textbf{42.65} & \textbf{61.48} & \textbf{24.61} \\
\modelentry{\includegraphics[width=7.5pt,height=7.5pt,keepaspectratio]{figures/logos/kimi.png}}{Kimi K3.1 Pro} & \underline{67.93} & \underline{27.95} & \underline{67.87} & \underline{32.51} & \underline{47.58} & 11.01 \\
\modelentry{\includegraphics[width=7.5pt,height=7.5pt,keepaspectratio]{figures/logos/xiaomi.png}}{MiMo-V2.6-Pro} & 56.89 & 19.99 & 55.72 & 29.24 & 36.71 & \underline{11.79} \\
\modelentry{\includegraphics[width=7.5pt,height=7.5pt,keepaspectratio]{figures/logos/zai.png}}{GLM-5.3-Flash} & 40.21 & 21.13 & 39.94 & 24.24 & 30.15 & 11.66 \\
\bottomrule
\end{tabular*}
\endgroup
\vspace{-16pt}
\end{table}

To analyze model performance across modalities, we compare IF and FP scores for VQA, AQA, and AVQA in Tab.~\ref{tab:modality_results}. Across all evaluated models, IF consistently decreases from VQA to AQA and further to AVQA, while AVQA also yields the lowest FP scores. For GPT-6 Astra, IF drops from 89.17\% on VQA to 71.68\% on AVQA, while FP decreases from 64.55\% to 42.01\%. Model strengths also vary across modalities, with Gemini 4 Argon leading IF on VQA and AVQA and Claude Sonnet 5.5 on AQA. For FP, GPT-6 Astra leads on VQA, Gemini 4 Argon on AQA, and Claude Opus 5.5 on AVQA. These results indicate that audio and cross-modal requirements are generally more challenging, particularly for instruction following, while different models exhibit distinct strengths across modalities.

\subsection{Tool Use and Efficiency Analysis}
\label{sec:tool_efficiency}

\begin{figure}[t]
    \centering

    \includegraphics[width=\columnwidth]{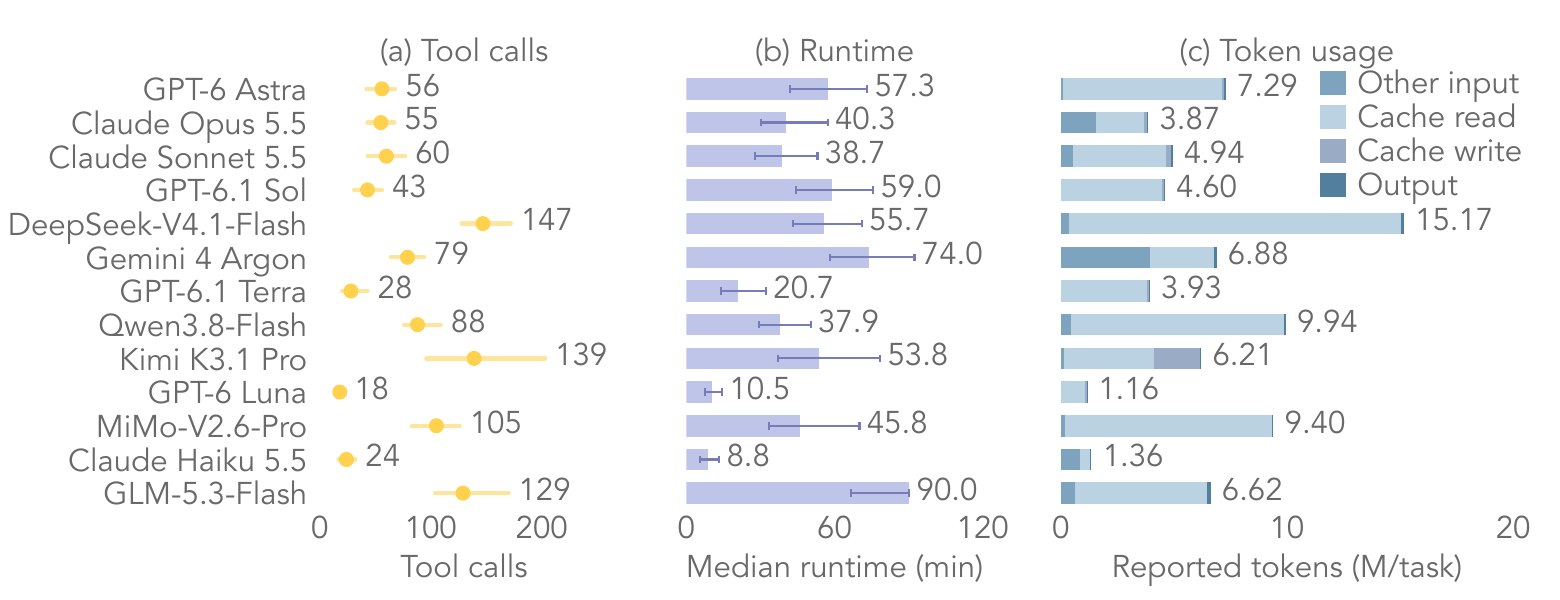}
    \caption{Median tool calls and runtime with interquartile ranges, and mean reported token usage per task.}
    \label{fig:tool_usage}
\vspace{-12pt}
\end{figure}

\bpstart{Tool usage}
We compare median tool calls and runtime, along with mean reported token usage, in Fig.~\ref{fig:tool_usage}. DeepSeek-V4.1-Flash uses a median of 147 calls and mean reported token usage of 15.17M per task, versus 56 calls and 7.29M for GPT-6 Astra, yet scores 2.74 in overall audiovisual quality versus 2.99. GLM-5.3-Flash has medians of 129 calls and 90.0 minutes per task but scores only 1.43, suggesting that extensive tool use does not necessarily improve overall audiovisual quality. Cache reads account for a substantial share of DeepSeek and MiMo token usage, reflecting extensive reuse of cached context.

\begin{figure}[t]
    \centering

    \includegraphics[width=\columnwidth]{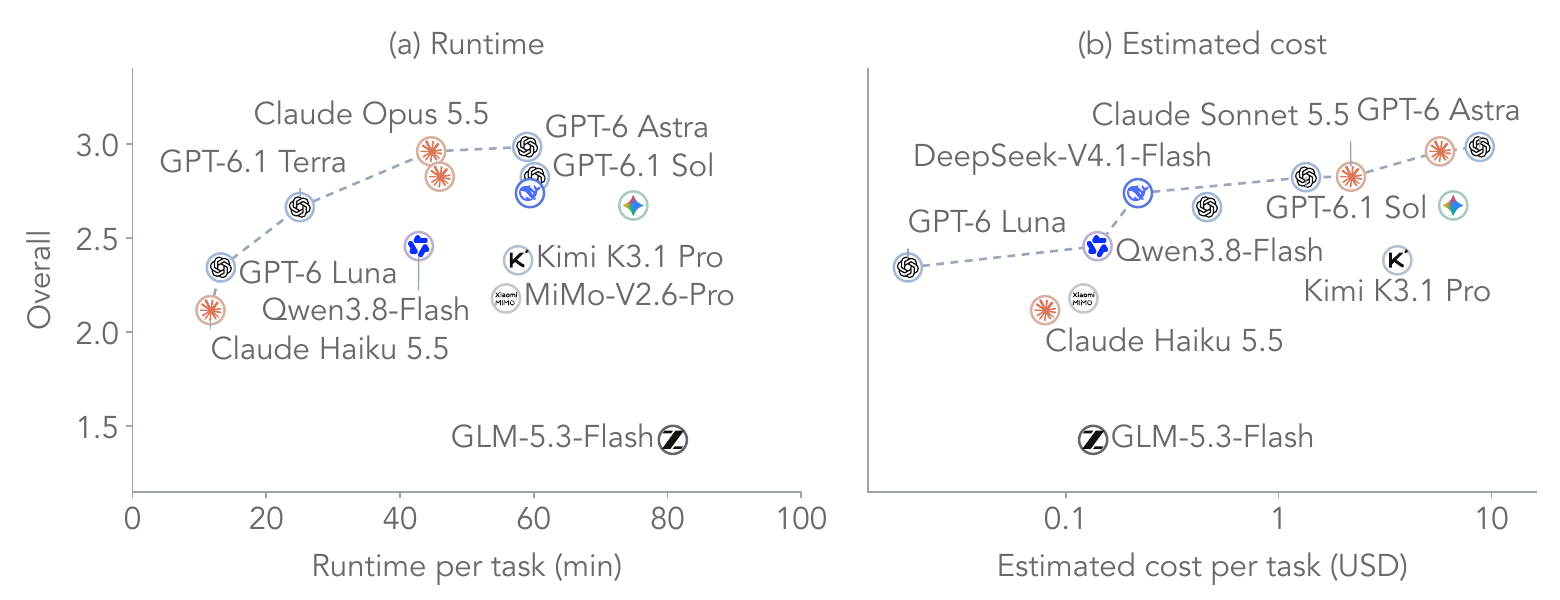}
    \caption{Overall audiovisual quality versus runtime and cost.}
    \label{fig:efficiency}
\vspace{-15pt}
\end{figure}

\bpstart{Runtime and cost}
We examine overall audiovisual quality with mean runtime and estimated cost in Fig.~\ref{fig:efficiency}. Claude Opus 5.5 scores 2.96, close to GPT-6 Astra's 2.99, while requiring shorter mean runtime and lower estimated cost. DeepSeek-V4.1-Flash achieves higher overall audiovisual quality than GPT-6.1 Terra at a lower estimated cost, whereas GLM-5.3-Flash combines the longest mean runtime with the lowest overall audiovisual quality, consistent with its frequent timeout failures shown in Fig.~\ref{fig:frameworks_failures}.

\subsection{Complexity Analysis}
\label{sec:complexity_analysis}

\begin{figure}[t]
    \centering

    \includegraphics[width=\columnwidth]{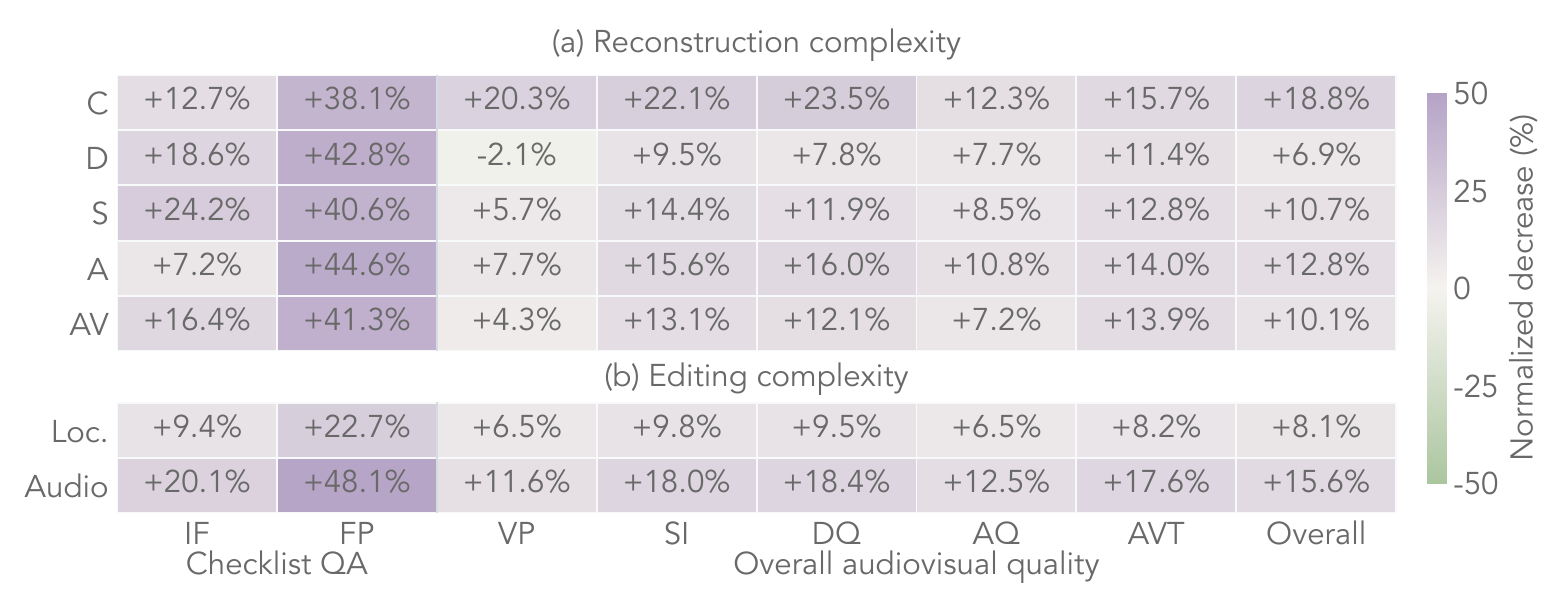}
    \caption{Normalized score gaps between low- and high-complexity groups, averaged across models (notation in Fig.~\ref{fig:data_statistics}).}
    \label{fig:complexity_analysis}
\vspace{-10pt}
\end{figure}

We compare scores between the lowest and highest complexity groups in Fig.~\ref{fig:complexity_analysis}. Across all five reconstruction complexity dimensions, FP shows the largest normalized gaps, ranging from 38.1\% to 44.6\% of its full scale. For audio reconstruction complexity, the normalized gap is 44.6\% for FP and 10.8\% for AQ, indicating a larger fidelity gap than audio-quality gap between these groups. Across all eight metrics, audio-source complexity shows larger gaps than object localization, including 48.1\% versus 22.7\% for FP and 15.6\% versus 8.1\% for overall audiovisual quality.

\subsection{Rendering Choices and Failure Analysis}
\label{sec:implementation_frameworks}

\begin{figure}[t]
    \centering

    \includegraphics[width=\columnwidth]{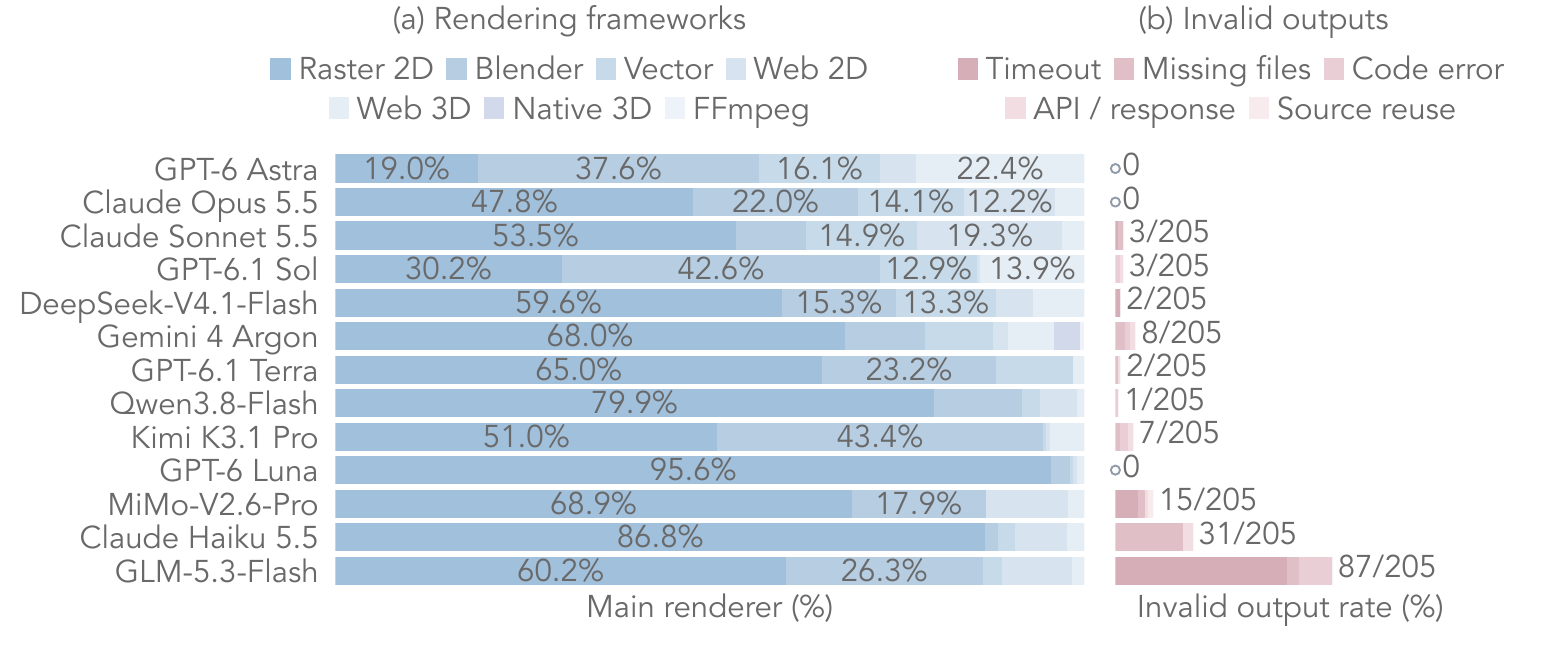}   
    \caption{Primary renderer among valid outputs (a) and invalid runs by cause out of 205 tasks per model (b).}
    \label{fig:frameworks_failures}
\vspace{-25pt}
\end{figure}

\bpstart{Rendering choices}
We analyze the rendering frameworks adopted by different models in Fig.~\ref{fig:frameworks_failures}. GPT-6 Astra and Claude Opus 5.5 achieve overall audiovisual quality scores of 2.99 and 2.96 with diverse renderers, including Blender, raster 2D, and Web 3D. In contrast, GPT-6 Luna and Claude Haiku 5.5 rely predominantly on raster 2D, accounting for 95.6\% and 86.8\% of their valid outputs, respectively, highlighting differences in rendering diversity between higher- and lower-performing models.

\bpstart{Failure analysis}
We examine execution failure patterns across models in Fig.~\ref{fig:frameworks_failures}, with timeouts accounting for the largest share at 81 runs, followed by missing required files at 44, code or runtime errors at 21, and API or response failures at 11. The latter include content filtering, server errors, authentication issues, and unusable responses. Two additional runs involve prohibited source reuse and are counted separately as compliance violations. Despite having no execution failures, GPT-6 Astra and GPT-6 Luna differ in FP, scoring 58.60\% and 23.01\%, respectively, so successful execution alone cannot ensure faithful reconstruction.

\subsection{Editing and Reconstruction Analysis}
\label{sec:editing_reconstruction}

We sample 50 tasks with three editing goals to test GPT-6.1 Sol, DeepSeek-V4.1-Flash, and Qwen3.8-Flash.

\begin{figure}[t]
    \centering

    \includegraphics[width=\columnwidth]{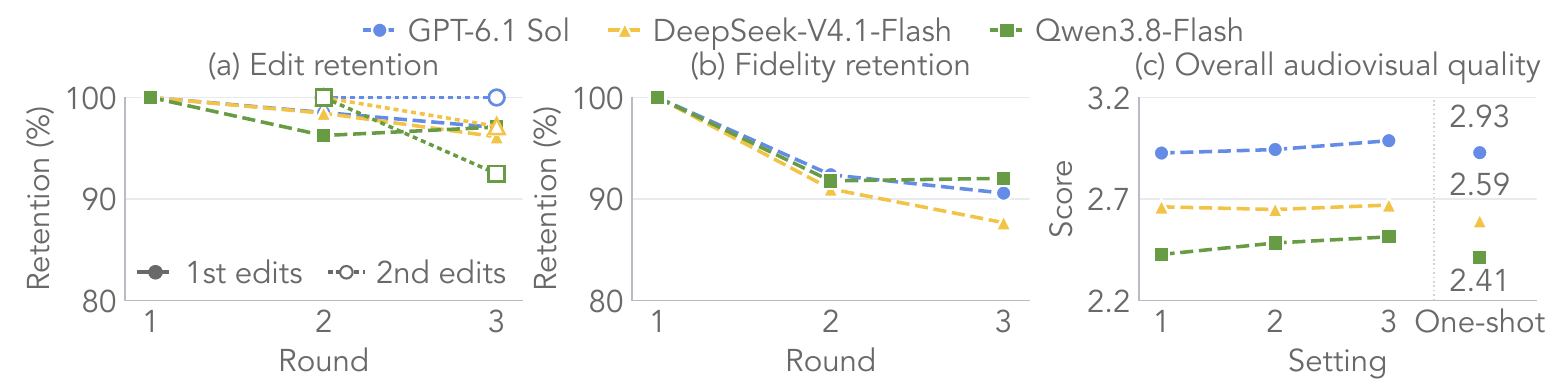}
    \caption{Edit retention, content preservation, and audiovisual quality across rounds versus three-goal one-shot editing.}
    \label{fig:multiround_comparison}
\vspace{-7pt}
\end{figure}

\bpstart{Multi-round editing}
We introduce one editing goal per round within the same agent session and code project to examine cumulative errors. As shown in Fig.~\ref{fig:multiround_comparison}, retention of previously passed edit checks remains above 92\%, while that of initially satisfied FP checks declines to 87.93--92.21\% by the third round. Overall audiovisual quality slightly exceeds our original one-shot setting, suggesting successive edits affect source preservation more than completed edits.

\begin{figure}[t]
    \centering

    \includegraphics[width=\columnwidth]{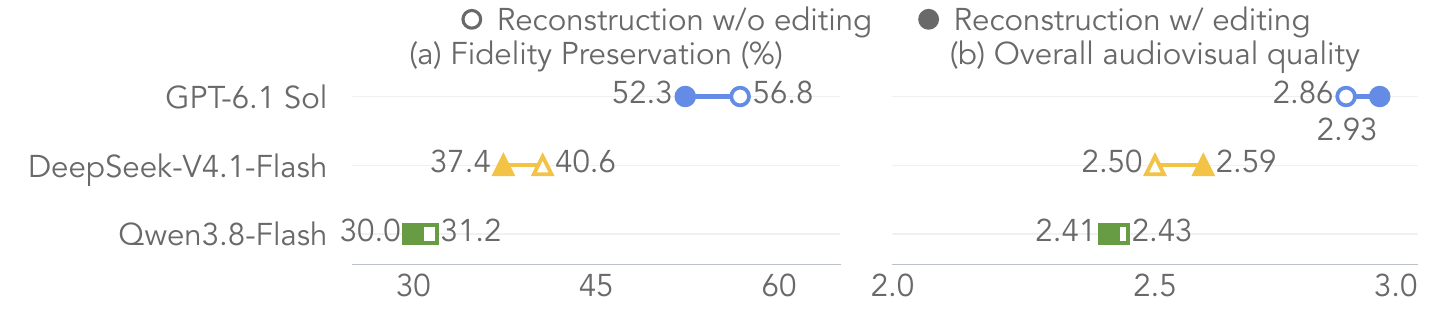}
    \caption{Reconstruction fidelity and overall audiovisual quality with and without editing.}
    \label{fig:reconstruction_comparison}
\vspace{-11pt}
\end{figure}

\bpstart{Impact of editing on reconstruction}
To isolate the challenges of editing, we introduce a reconstruction-only setting and compare it with editing using the same FP checklist in Fig.~\ref{fig:reconstruction_comparison}. Both FP and audiovisual quality are low without editing, especially for weaker models. Editing adds preservation challenges, lowering FP across all models, with larger declines but gains in audiovisual quality for stronger models.

\subsection{MLLM Judge Validation}
\label{sec:judge_validation}

\begin{table}[t]
\centering

\caption{Judge validation. Pair reports human--judge preference agreement on non-tied pairs, Model uses Spearman's $\rho$ between human and judge model rankings, Stability uses ICC(A,3) for Argon's three-run mean, and Cross-Judge uses video-level Cronbach's $\alpha$.}
\label{tab:judge_validation}
\begingroup
\footnotesize
\setlength{\tabcolsep}{1pt}
\renewcommand{\arraystretch}{1.1}
\begin{tabular*}{\columnwidth}{@{\extracolsep{\fill}}lcccc@{}}
\toprule
\textbf{Metric}
& \textbf{Pair}
& \textbf{Model}
& \textbf{Stability}
& \textbf{Cross-Judge} \\
\midrule
Instruction Following & 0.976 & 0.917 & 0.959 & 0.862 \\
Fidelity Preservation & 0.960 & 0.933 & 0.992 & 0.921 \\
Overall Audiovisual Quality & 0.980 & 0.850 & 0.976 & 0.871 \\
\bottomrule
\end{tabular*}
\endgroup
\vspace{-15pt}
\end{table}

We recruit three human reviewers to independently compare 205 video pairs from nine models under blinded conditions with institutional ethics approval, while Gemini 4 Argon scores all 410 videos in three independent runs for repeat stability. Human alignment uses pairwise agreement and correlations between Bradley--Terry human rankings and judge rankings based on model means, with Kendall's $W$ across reviewers' rankings reaching 0.937 for IF, 0.800 for FP, and 0.826 for overall audiovisual quality.
Consistency among Gemini 4 Argon, Gemini 3.8 Flash, and MiMo-V2.6-Pro is examined using video-level Cronbach's $\alpha$ and nine-model rank correlations, with pairwise Spearman correlations ranging from 0.783 to 0.950 for IF, 0.950 to 0.983 for FP, and 0.867 to 0.983 for overall audiovisual quality, indicating strong ranking agreement. Tab.~\ref{tab:judge_validation} summarizes the results.

\section{Conclusion}
We introduce \benchmark{}, a benchmark for agentic code-based audio-video editing with 1,026 editing tasks and 38 editing operations. Our dual-level evaluation measures edit correctness, reconstruction fidelity, and audiovisual quality. Evaluations of 13 model configurations reveal persistent challenges in editing and reconstruction, particularly for audio and cross-modal requirements. We believe that \benchmark{} will help advance multimodal understanding, temporal reasoning, and reliable agentic editing, paving the way for more capable and controllable audio-video creation.

\section*{Impact Statement}

AVCodeEdit aims to advance the evaluation of coding agents for audiovisual reconstruction and editing.
Code-based editing may facilitate reproducible and customizable media creation in education, design, and creative production.
Such capabilities may also be misused to create misleading audiovisual content.
The dataset release consists of source video links and task annotations. Source videos remain with their original hosts.
Our human evaluation was approved by our institution's ethics review board, and all participants gave informed consent.

\section*{AI Disclosure Statement}

GPT-6.1 Sol and Gemini 3.8 Flash assisted with dataset construction, editing-instruction development, and checklist preparation, as described in the paper.
Gemini 4 Argon served as the primary evaluation judge.
ChatGPT assisted with manuscript preparation and code development.
Research ideas and conclusions were developed by the authors, who take full responsibility for all submitted content.

\bibliographystyle{icml2027}
\bibliography{references}
\end{document}